\documentclass{article}

\usepackage[utf8]{inputenc}
\usepackage{amsmath,amssymb,amsthm}
\usepackage{hyperref}
\usepackage[inline]{enumitem}
\usepackage{geometry}
\usepackage{graphicx}
\usepackage{booktabs}
\usepackage{multirow}
\usepackage{tikz}
\usepackage{wrapfig}
\usetikzlibrary{calc}
\newtheorem{lemma}{Lemma}

\newtheorem{theorem}[lemma]{Theorem}

\title{Balancing Privacy, Robustness, and Efficiency \\ in  Machine Learning}
\author{%
Youssef Allouah \and Rachid Guerraoui \and John Stephan
}
\date{EPFL, Switzerland}

\begin{document}

\maketitle

\begin{abstract}
This position paper argues that achieving robustness, privacy, and efficiency simultaneously in machine learning systems is infeasible under prevailing threat models. The tension between these goals arises not from algorithmic shortcomings but from structural limitations imposed by worst-case adversarial assumptions. We advocate for a systematic research agenda aimed at formalizing the robustness-privacy-efficiency trilemma, exploring how principled relaxations of threat models can unlock better trade-offs, and designing benchmarks that expose rather than obscure the compromises made. By shifting focus from aspirational universal guarantees to context-aware system design, the machine learning community can build models that are truly appropriate for real-world deployment.
\end{abstract}

\newcommand{\efficiency}{efficiency}
\newcommand{\thesisbox}[1]{%
  \begin{center}\vspace{-0.5em}%
    \fbox{\parbox{0.93\linewidth}{\textbf{Position:}~#1}}%
  \end{center}\vspace{-0.8em}}

\section{Introduction}
Machine learning systems now underpin high-stakes applications, such as cancer diagnosis in radiology, legal document automation, and predictive keyboards used daily by billions of people, dramatically increasing the stakes of privacy violations and adversarial manipulations.
Each deployment is expected to protect \emph{user data}, resist \emph{malicious manipulation}, and run at \emph{large scale}.  Meeting all three demands simultaneously has proven elusive.  Wan et al.~\cite{wan2023poisoning} showed that as few as 100 poisoned instruction‑tuning examples can reliably subvert GPT‑style models across hundreds of tasks.  On the edge‑device side, Bagdasaryan et al.~\cite{bagdasaryan2020backdoor} demonstrated that only two compromised phones can implant a hidden back‑door in a federated keyboard predictor equipped with secure aggregation and client‑side differential privacy; Suliman and Leith~\cite{suliman2023two} later reproduced the attack against Google GBoard at production scale.  These findings suggest a broader structural tension: essentially, strengthening one pillar of \emph{robustness}, \emph{privacy}, or \emph{\efficiency{}} often compromises the remaining two.

\textbf{Limitations of worst-case models.}
Many existing analyses default to the strongest  adversary assumptions: fully malicious clients, local differential privacy without any trusted intermediary, and full collusion among attackers.  While these worst-case models yield clean theoretical guarantees, they often overshoot practical needs and obscure realistic trade-offs.  In many applications, adversaries have limited coordination, semi-honest data shufflers exist, or partial trust anchors can be leveraged.  Recognizing where worst-case assumptions are unnecessarily pessimistic is key to designing systems that recover valuable efficiency or robustness without materially raising actual risk.

\vspace{2mm}

\thesisbox{Under conventional \emph{worst‑case} threat models (adversaries), a machine learning system cannot simultaneously satisfy robustness to corruption, user‑level privacy, and system‑wide \efficiency{}.
Rather than pursuing an unattainable ideal, we advocate explicit and context‑dependent compromises with transparent reporting on all three axes.}
\vspace{2mm}

\paragraph{Why now?}
Foundation models amplify the stakes: a single poisoned prompt, once assimilated during  fine‑tuning, can ripple across thousands of downstream applications~\cite{wan2023poisoning}; conversely, stronger integrity checks can multiply training cost by orders of magnitude.  At the same time, rapidly advancing regulations—such as Europe's AI Act~\cite{euai2023} which explicitly mandates transparency in robustness and privacy trade-offs, and increasingly stringent healthcare and financial-sector regulations requiring demonstrable guarantees—are pushing these tensions urgently into practical focus. Practitioners sometimes ignore the tension, ending up with solutions that do not scale, are not robust or not private. Sometimes they are forced to choose which requirement to relax, often without guidance on the systemic implications.  A principled framework for reasoning about this three‑way trade‑off is urgently needed.

\paragraph{Contributions.}
We summarize our contributions as follows:
\begin{enumerate}
    \item \textit{Synthesis of evidence:} A unified statement of the robustness–privacy–\efficiency{} trilemma, drawing on both theoretical lower bounds and empirical failures.
    \item \textit{Mapping of deployments:} A trilemma simplex diagram placing four representative systems, namely healthcare collaboration, autonomous fleets, consumer keyboards, and user-specific LLM fine-tuning, into explicit trade-off regions.
    \item \textit{Design guidance:} A set of practical guidelines, including suggestions for open-source benchmarking to report privacy, robustness, and \efficiency{} jointly.
    \item \textit{Research agenda:} An articulation of open problems around threat-model relaxations, adaptive privacy budgets, and certifiable trade-off disclosure.
\end{enumerate}

\paragraph{Related work.}
Our trilemma \emph{complements} a growing body of “can’t-have-it-all’’ results.
El Mhamdi et al.~\cite{el2022impossible} argued that large ML models cannot be simultaneously accurate, private, and tamper-resilient without unrealistic assumptions on data and compute.
In contrast, we highlight efficiency  as a fundamental third axis interacting with privacy and robustness,  map the corresponding trilemma onto real deployments (Figure~\ref{fig:triangle}), and advocate for standardized disclosure practices to transparently communicate these trade-offs.
Recently, in large language model applications, Rando \& Tramèr~\cite{randouniversal} demonstrated that reinforcement learning from human feedback remains vulnerable to back-door insertion.
For federated learning with heterogeneous data, Karimireddy et al.~\cite{karimireddy2021byzantine} prove that any protocol tolerating $\eta$-fraction of malicious clients incurs an $\Omega(\eta)$ accuracy loss even without privacy noise.
Most recently, Xie et al.~\cite{xie2023unraveling} formally linked DP budgets to certified robustness radii in federated learning, confirming that boosting one shrinks the other.

Over the last decade, researchers have explored the interaction between privacy and robustness. Dwork \& Lei~\cite{dwork2009differential} showed that classical robust estimators admit differentially private versions. Ma et al.~\cite{ma2019data} then quantified precisely how many poisons a DP learner can tolerate, and Cheu et al.~\cite{cheu2021manipulation} revealed that even tiny coalitions can fully subvert local-DP protocols. Naseri et al.~\cite{naseri2022flRobustDP} empirically found that DP noise blocks some FL backdoors while leaving others intact. Asi et al.~\cite{asi2023robustness} provided a black-box DP-robustness reduction, and Allouah et al.~\cite{allouah2023privacy,allouah2025towards} formalized a privacy–robustness trade-off and introduced the CAF aggregator to hit its optimal point efficiently using secure shared randomness. These milestones underpin our trilemma framework and its design guidance.

\section{The Robustness--Privacy--Efficiency Trilemma}
\label{sec:trilemma-evidence}
We focus on three system‑level properties that every large‑scale ML deployment must balance ideally.
Table~\ref{tab:properties} summarizes how each property is commonly quantified in practice.
We explicitly distinguish the server and users in our terminology given our focus on privacy, although our discussion encapsulates  classical centralized ML systems when trusting the server.

\paragraph{Objectives.}
We focus on three system-level properties essential for trustworthy deployment of large-scale ML systems: \emph{privacy}, \emph{robustness}, and \emph{efficiency}. Privacy involves formally quantifiable guarantees, such as differential privacy~\cite{dwork2014algorithmic}, limiting information leakage about individual users from communicated model updates or outputs.
Robustness ensures that systems maintain integrity and accuracy despite malicious behaviors such as adversarial participants or poisoned data. Lastly, efficiency encompasses the feasibility of training and deploying models at large scales, capturing aspects of computational, communication, and statistical efficiency.

\paragraph{Tools.}
A wide range of practical techniques has been developed to meet these objectives, each presenting its own trade-offs between privacy, robustness, and efficiency.
As one of the most established privacy frameworks, differential privacy (DP)~\cite{dwork2014algorithmic} provides information-theoretic  guarantees of statistical indistinguishability, ensuring protection even against adversaries with unbounded computational power.
It is typically implemented by injecting noise, most commonly Gaussian or Laplacian, into user updates or aggregated statistics.
Although DP mechanisms are relatively simple and computationally efficient, the added noise inevitably leads to utility loss~\cite{duchi2013local,bassily2014private}, especially in large-scale or distributed machine learning settings.
Still, DP offers some of the most rigorous and durable privacy guarantees available in practice: once applied correctly, it ensures that the released model remains private, even if it is later shared with third parties, made publicly available, or subjected to arbitrary post-processing~\cite{dwork2014algorithmic}.
In contrast, cryptographic approaches such as homomorphic encryption (HE) and secure multi-party computation (MPC) can avoid the need for noise injection by assuming a more realistic, computationally bounded adversary, although their goal is not to protect the output, but rather intermediate communications. Yet, as discussed in Section~\ref{sec_he}, these methods may be used to improve utility under DP, and come with their own set of limitations.

On the robustness front, a variety of robust aggregation schemes have been proposed to defend against adversarial participants or corrupted data.
Simple yet effective methods such as the coordinate-wise median and trimmed mean~\cite{yin2018byzantine,allouah2023fixing,dahanfault} replace the standard average with robust estimators that limit the influence of outliers.
More sophisticated techniques like the geometric median~\cite{cohen2016geometric}, adaptive clipping~\cite{allouah2025adaptive}, and spectral filtering~\cite{diakonikolas2019sever,kamath2024broader} further enhance robustness by leveraging global structure or adaptively bounding update magnitudes.
These methods collectively aim to maintain model integrity even in the presence of malicious clients, offering strong adversarial resilience at only moderate computational cost~\cite{dong2019quantum,karimireddy2021byzantine,allouah2023fixing}.

\begin{table}[t]
\centering\small
\begin{tabular}{@{}p{0.15\linewidth}p{0.25\linewidth}p{0.50\linewidth}@{}}
\toprule
\textbf{Axis} & \textbf{Level} & \textbf{Typical realization / tooling}\\
\midrule
\multirow{4}{=}{\textbf{Privacy}} 
 & \emph{Trusted server} (\textit{central} DP) & DP-SGD; per-round noise at server\\
 & \emph{Trusted shuffler} & Privacy amplification by shuffling\\
 & \emph{Secure shared randomness} & Correlated-noise protocols; pairwise masks\\
 & \emph{No trust} (\textit{local} DP) & Randomized response; LDP-FedAvg\\
\midrule
\multirow{4}{=}{\textbf{Robustness}} 
 & Data poisoning & Dirty-label, feature collision\\
 & Model poisoning & Arbitrary gradient submission; Sybil collusion\\
  & Backdoor/trigger & Model-replacement attack; controlled ASR\\
\midrule
\multirow{3}{=}{\textbf{Efficiency}} 
 & Communication cost & Bits-per-client-per-round; gradient sparsification\\
 & Computation cost & Per-round FLOPs; HE or MPC overhead\\
 & Statistical efficiency & Num. of samples to reach target loss; heterogeneity cost\\
\bottomrule
\end{tabular}
\vspace{1mm}
\caption{Selectable trust/attack levels on each axis.  We recommend this type of taxonomy to state exactly which trust levels ML systems occupy.}
\label{tab:properties}
\end{table}

\subsection{Theoretical Evidence}
We motivate the trilemma through a representative impossibility result.
We consider $n$ users, each with a local dataset, aiming at learning a model collaboratively using some algorithm $\mathcal{A}$.
We choose distributed mean estimation as a canonical representative task because many more complex machine learning procedures, including  SGD~\cite{bertsekas2015parallel} for deep learning models, reduce to mean estimation under standard convexity or smoothness assumptions~\cite{el2021collaborative,el2022impossible}.
We consider that algorithm $\mathcal{A}$  must satisfy local differential privacy (LDP)~\cite{kasiviswanathan2011can}, the strongest DP requirement whereby all user communications are DP and no entity is trusted for privacy.
Algorithm $\mathcal{A}$ must also be robust to $\eta$-corruption; $\mathcal{A}$ should guarantee low estimation error despite an unknown $\eta$-fraction of users arbitrarily deviating from the algorithm, including using corrupt data.
This is similar to the strongest adversary model in statistics; namely Huber's contamination model~\cite{huber1992robust}, or its equivalent in distributed systems~\cite{lamport82Byzantine}.

Below, we paraphrase a result due to Allouah et al.~\cite{allouah2023privacy} on a privacy-robustness lower bound in the above threat model.
We briefly recall that $\varepsilon$ is the LDP budget, $\eta$ the maximal corruption fraction, $d$ the model dimension, and $G^2$ the data heterogeneity; here, the variance across honest user inputs.

\begin{theorem}[Privacy-robustness lower bound {\cite{allouah2023privacy}}, paraphrased]
\label{thm:pr-robust}
Consider distributed mean estimation in $\mathbb{R}^{d}$ with $n$ users, of which an $\eta$‑fraction may behave maliciously.  
Any algorithm, producing an estimate $\hat{\mu}$ of the true mean $\mu \in  \mathbb{R}^d$ of inputs, and achieving $(\varepsilon,\delta)$‑local differentially privacy  and robustness to $\eta$‑corruption must incur an error 
\begin{equation}
\mathbb{E}\bigl\|\hat{\mu}-\mu\bigr\|_2^2
=
\widetilde{\Omega}\left(\frac{\eta}{\varepsilon^2} + \frac{d}{\varepsilon^2 n} + \eta G^2\right).
\end{equation}
\end{theorem}

The above result shows that, regardless of algorithmic design, there is a fundamental cost, in terms of error or accuracy, to ensuring the strongest form of both DP and robustness to corruption.
The first term is of special interest; it quantifies an interaction between robustness and privacy as a cross term involving the corruption fraction $\eta$ and the privacy budget $\varepsilon$.
Precisely, at a fixed target error, the maximal corruption fraction $\eta$ must decrease quadratically with the privacy budget $\varepsilon$.
The other two terms quantify the \emph{independent} fundamental costs of local DP~\cite{duchi2013local} and robustness to corruption~\cite{lai2016agnostic,karimireddy2021byzantine}.
Finally, note that the lower bound is tight for strongly convex tasks~\cite{allouah2023privacy}, and also extends to discrete distribution estimation~\cite{cheu2021manipulation,acharya2021robust}.

\paragraph{Efficiency as a fundamental third pillar.}
While Theorem~\ref{thm:pr-robust} formalizes the inherent trade-off between DP and robustness, we argue that an equally fundamental dimension emerges in practice: computational and communication efficiency. This third axis forms a trilemma with privacy and robustness, where jointly optimizing for all three objectives represents a major challenge.

Despite the theoretical utility limitations of combining DP with robustness, Allouah et al.~\cite{allouah2023privacy} proposed the \textsc{SMEA} aggregation scheme, which matches the lower bound in Theorem~\ref{thm:pr-robust} and achieves the lowest known error among robust aggregation methods under local DP constraints. However, this utility gain comes at a substantial computational cost: \textsc{SMEA} has a runtime complexity of $\widetilde{\mathcal{O}}\left( \exp(\eta \cdot  n)  d^3 \right)$, scaling exponentially with the number of tolerated adversaries $\eta \cdot n$ and cubically with the model dimension $d$.
For instance, in ImageNet-scale models~\cite{deng2009imagenet} where $d\!>\!10^7$, such computational costs quickly become prohibitive, especially in large-scale federated learning (FL) systems.
Indeed, due to this computational burden, empirical evaluation was limited to a small logistic regression task ($d=69$) and a modest number of users ($n=7$)~\cite{allouah2023privacy}.
To mitigate these scalability issues, the CAF aggregator~\cite{allouah2025towards} was introduced as a polynomial-time alternative with complexity $\mathcal{O}(\eta \cdot n \cdot d^3)$, maintaining theoretical optimality while improving tractability. While CAF reduces the exponential dependence on $\eta \cdot n$ thanks to ideas from high-dimensional robust statistics~\cite{diakonikolas2019robust}, its cubic scaling in $d$ remains a bottleneck for contemporary high-dimensional models~\cite{openai2023gpt4, touvron2023llama}. Nevertheless,  approaches similar to~\cite{allouah2025towards} offer a promising pathway for navigating the trilemma under relaxed adversarial models, as further discussed in Section~\ref{sec_dp_relax}.
This exemplifies the core challenge faced when ensuring privacy and robustness in practice: one must choose between conventional robust aggregation methods~\cite{yin2018byzantine,karimireddy2021byzantine}—efficient but significantly degraded by DP noise (e.g., see Figure 2 in~\cite{allouah2025towards})—and theoretically optimal schemes~\cite{allouah2023privacy, allouah2025towards}, which achieve better utility at the cost of prohibitive computation in practice.

Finally, communication efficiency emerges as a distinct yet closely interdependent aspect of efficiency in the trilemma.
The lower bound in Theorem~\ref{thm:pr-robust} includes the term $\tfrac{d}{\varepsilon^{2}n}$, which directly ties model dimensionality $d$ to both communication cost and achievable utility. As $d$ increases, the user communication load grows proportionally, and the utility decreases, unless one relaxes privacy (i.e., increases $\varepsilon$) or scales up the number of participants $n$. Recent work on privacy-preserving scaling laws~\cite{mckenna2025scaling} reinforces this trade-off, showing that to maintain constant utility as model complexity increases, one may be forced to weaken privacy guarantees. We further explore the theoretical trade-offs between communication, privacy, and utility in Appendix~\ref{sec_priv_comm}.

\subsection{Empirical Evidence}

The theoretical limitations associated with simultaneously enforcing DP and robustness, such as those captured by the lower bound in Theorem~\ref{thm:pr-robust}, are strongly reflected in empirical results across the literature.
Notably, most works on robust FL~\cite{yin2018byzantine, baruch2019alittle,Karimireddy2021History,shejwalkar2022back,farhadkhani2022byzantine,allouah2023fixing,gorbunov2023variance}, even without privacy constraints, are typically evaluated only on simple ML tasks such as MNIST~\cite{mnist} and CIFAR-10~\cite{cifar}. While these datasets are considered simple in standard, non-adversarial settings, they become substantially more challenging under adversarial conditions.

The difficulty is compounded when privacy is introduced into the learning process through DP noise injection. In such cases, existing approaches struggle to scale beyond basic datasets. For example, several recent works~\cite{guerraoui2021differential,allouah2023privacy} are constrained to extremely simple ML tasks like small-scale logistic regression, while others are unable to go beyond Fashion-MNIST~\cite{allouah2025towards}.
In addition to mirroring the theoretical trade-offs, these empirical limitations also cast doubt on the performance and scalability of privacy-preserving and robust solutions when applied to contemporary ML tasks, characterized by massive models~\cite{openai2023gpt4,touvron2023llama}, high-dimensional data~\cite{deng2009imagenet,lin2014microsoft}, and complex multimodal objectives~\cite{radford2021learning}, far beyond the simplicity of benchmarks like MNIST or CIFAR-10.
This widening gap between theoretical promise and empirical scalability presents a critical barrier to achieving privacy and robustness in realistic environments.
To provide a broader representative view of the practical cost of privacy or robustness in large-scale systems, we report in Table~\ref{tab:empirical} the results of three recent studies spanning edge,
cloud, and centralized settings.

\begin{table}[ht!]
\centering
\small
\begin{tabular}{p{0.18\linewidth}p{0.28\linewidth}p{0.15\linewidth}p{0.08\linewidth}p{0.15\linewidth}}
\toprule
\textbf{Scenario} & \textbf{Model \& data} & \textbf{Attack} & \textbf{$\eta$} & \textbf{Outcome}\\
\midrule
Mobile keyboard FL \cite{bagdasaryan2020backdoor} & 20k clients, next‑word pred. & Trigger back‑door & 0.01 \% & 94\% trigger accuracy\\[0.15em]
LLM Instruction tuning \cite{wan2023poisoning} & 1 client (central), LLaMA‑7B, Stack‑OF & Single‑sent. poison & 1\% & 92\% task failure\\[0.15em]
Fashion-MNIST FL \cite{allouah2025towards} & 100 clients, CNN & Model poison & 10\% & 80\% attack accuracy \\[0.15em]
\bottomrule
\end{tabular}
\vspace{1mm}
\caption{Representative empirical demonstrations of the trilemma.  
Even minute malicious participation or single‑example poisoning circumvents privacy
noise and overwhelms lightweight integrity checks; conversely, stronger
defenses would breach communication or compute budgets.}
\label{tab:empirical}
\end{table}

\section{Case Studies: Context-Specific Trade-offs}
We highlight four illustrative scenarios; namely healthcare collaboration, autonomous vehicle fleets, consumer keyboards, and personalized language-model fine-tuning; chosen for their diversity, broad practical relevance, and explicit representation of distinct points in the robustness–privacy–efficiency trade-off landscape.
Figure~\ref{fig:triangle} qualitatively positions four representative deployments inside the
robustness–privacy–\efficiency{} simplex according to publicly documented
design choices and performance budgets.
The pattern is consistent: every deployment hugs the line joining its two
chosen priorities and pays a tangible cost on the third.

\begin{figure}[t]
\centering
\begin{tikzpicture}[scale=2.5]
  \coordinate (R) at (0,1);
  \coordinate (P) at (-0.866,-0.5);
  \coordinate (S) at (0.866,-0.5);
  \draw[thick] (R)--(P)--(S)--cycle;
  \node[above]       at (R) {\textbf{Robustness}};
  \node[below left]  at (P) {\textbf{Privacy}};
  \node[below right] at (S) {\textbf{Efficiency}};

  \node[fill=blue!60 ,circle,inner sep=1.4pt,
        label=above left:{\scriptsize Healthcare FL}]         at ($(P)!0.75!(S)$) {};
  \node[fill=red!70  ,circle,inner sep=1.4pt,
        label=above right:{\scriptsize Autonomous fleets}]          at ($(R)!0.65!(S)$) {};
  \node[fill=green!70,circle,inner sep=1.4pt,
        label=below:{\scriptsize Consumer keyboards}]         at ($(P)!0.25!(S)$) {};
  \node[fill=orange!80,circle,inner sep=1.4pt,
        label=above left:{\scriptsize Personal LLM tuning}]  at ($(R)!0.35!(P)$) {};
\end{tikzpicture}
\caption{Qualitative placement of four production‑motivated systems in the
robustness–privacy–\efficiency{} trilemma.  Each dot lies nearer to the two
properties it explicitly optimizes and away from the sacrificed one. For instance, Healthcare FL is positioned towards privacy and efficiency due to low epsilon DP in federated learning across numerous trusted hospitals~\cite{rieke2020future}.}
\label{fig:triangle}
\end{figure}

\subsection{Healthcare Machine Learning}
Multi‑institution consortia train diagnostic models on X‑ray or MRI data using
\textsc{FedAvg}~\cite{mcmahan17a} with secure aggregation and client‑side DP in
order to comply with HIPAA/GDPR while limiting inter‑hospital bandwidth~\cite{sheller2020federated,rieke2020future}.  These systems therefore prioritize
\emph{privacy} (privacy budgets, e.g., $\varepsilon\!\approx\!1$–$3$) and
\emph{\efficiency{}} (20–100 hospitals, training times near the centralized
baseline).  Robustness is the deprioritized corner: experimental work shows
that poisoning only 10\% of hospitals can flip a pneumonia classifier from
85\% to below chance accuracy without detection~\cite{alkhunaizi2022suppressing}, a failure mode exacerbated by the DP noise that
masks outliers and prevents robust aggregation.

\subsection{Autonomous Fleets}
Autonomous vehicles and drone swarms exchange model updates via low‑latency
V2X links to adapt to traffic and sensor drift~\cite{nguyen2021federated}.  Designers therefore prioritize
\emph{robustness}—so that a small set of compromised vehicles cannot trigger
cascading failures—and real‑time \emph{\efficiency{}}, keeping round‑trip
update latency below \,100ms through lightweight message formats and
sparsified gradients.  Privacy is the relaxed corner: DP mechanisms would introduce latency or accuracy degradation unacceptable for real-time control tasks, and cryptographic secure aggregation incurs prohibitive computational overheads given sub-second latency requirements in vehicular networks.

\subsection{Consumer Keyboards and Recommenders}
Mobile keyboards, next‑word predictors, and on‑device recommenders serve
hundreds of millions of users daily.  Commercial deployments such as Google
GBoard train language models with \textsc{FedAvg} augmented by
secure aggregation and per‑client DP budgets, e.g., $\varepsilon<10$, to respect user data regulations while keeping
per‑round uplink traffic below 40kB
\cite{hard2018federated,kairouz2021advances}.  These choices elevate
\emph{privacy} and \emph{\efficiency{}}: training converges within a few
hours of wall‑clock time across millions of devices.  Robustness, however, is unaddressed; Bagdasaryan et al.\ \cite{bagdasaryan2020backdoor} show
that only two malicious phones (\( \eta <\!0.01\%\)) can inject a
trigger that forces the model to suggest attacker‑chosen words with
94\% success, the DP noise having obscured the anomalous gradients from server‑side detection.

\subsection{Personal Large‑Language‑Model Fine‑Tuning}
Software as a service (SaaS) platforms increasingly let end‑users upload private corpora—e‑mails,
medical notes, legal templates—to personalize a foundation model via
parameter‑efficient adapters, e.g., LoRA~\cite{hu2022lora}, p‑tuning~\cite{li2021prefix}.  
Because the raw text never leaves the provider’s enclave, such services
foreground \emph{privacy}.  
To curb prompt‑injection and jailbreak risks, providers run lightweight
validation of the user‑supplied gradients or adapters against a trusted
reference set, giving a partial degree of \emph{robustness}.  
Efficiency is the loosest corner: storing and periodically re‑validating
millions of per‑user adapter matrices incurs a linear memory cost
(MB$\times$users) and adds seconds of latency to each deployment update cycle.  
Even then, Wan et al.~\cite{wan2023poisoning} show that a single poisoned
instruction suffices to steer model behavior across hundreds of downstream
tasks, underscoring the residual vulnerability when robustness is treated as
secondary.

\subsection{Take‑away}
Across healthcare, autonomous fleets, consumer mobile services, and personal LLM fine-tuning, a consistent pattern emerges: every deployment explicitly optimizes at most two of the three system-level properties, while the third property, though still desirable, often need not be maximized under its conventional worst-case adversary model.
However, it remains important to quantify the exact amount of relaxation on that least-prioritized aspect.
Those that prioritize privacy and \efficiency{} expose themselves to
malicious manipulation; those that protect against attackers either forego user‑level privacy or incur prohibitive cost.  
This empirical tour therefore reinforces the theoretical limits of
Section~\ref{sec:trilemma-evidence} and motivates the question we tackle next:
\emph{how should practitioners navigate the unavoidable trade‑offs when
designing trustworthy ML systems?}

\section{Design Implications}
\label{sec:design}
The case studies confirm that the trilemma is not only academic; many real systems already relax one requirement out of the three we focus on. We distill four practices that help designers navigate these trade‑offs consciously.

\paragraph{Context-specific system design.}
Machine learning systems should prioritize two of the three axes—robustness, privacy, and efficiency— while precisely relaxing the third axis, based on domain-specific threat models and operational constraints. In safety-critical settings, such as autonomous vehicles, robustness must dominate even at the cost of some privacy guarantees. In personal applications involving sensitive user data, such as healthcare or private assistants, privacy may take precedence, accepting weaker robustness guarantees against rare adversaries. Designing systems without an explicit prioritization inevitably leads to brittle or illusory security properties.

\paragraph{Transparent joint benchmarking.}
Reporting only a single security metric hides the true cost. A joint benchmark
should report, for every defense:
\begin{itemize}
    \setlength{\itemindent}{0pt}
    \setlength{\leftmargini}{0pt}
    \item \textit{Privacy:} $(\varepsilon,\delta)$-DP budget and membership inference attack success~\cite{shokri2017membership}.
    \item \textit{Robustness:} malicious fraction $\eta$ or attack success rate under a stated threat model.
    \item \textit{Efficiency:}  wall‑clock training time, per‑client
    bytes/round, and the accuracy gap to a no‑defense baseline, in the spirit of the MLPerf methodology~\cite{reddi2020mlperf}.
\end{itemize}
When all three columns sit in the same table, corner‑specific costs become
hard to bury.
For example, a federated language model might transparently report:
\begin{enumerate*}
  \item \textit{Privacy}: $(\varepsilon=10, \delta=10^{-6})$ and 5\% membership inference accuracy.
  \item \textit{Robustness}: Tolerates 1\% malicious participation with at most 10\% accuracy drop.
  \item \textit{Efficiency}: 40 kB/client/round, training completed within 4 hours wall-clock, accuracy gap $<2\%$ compared to non-private baseline.
\end{enumerate*}

\paragraph{Explicit threat model disclosure.}
Security claims hinge on assumptions that are too often implicit.
We recommend a five‑field template mirroring the new NeurIPS reproducibility
checklist item on threat models:
\textit{(i) adversary capability, (ii) collusion level, (iii) trusted entities,
(iv) acceptable failure modes, (v) evaluation surface}. Using this template
prevents apples‑to‑oranges comparisons and deters over‑stated guarantees; a
comprehensive taxonomy appears in the survey by
Mothukuri et al.~\cite{mothukuri2021survey}.
For example, a threat-model disclosure might clearly state:
\begin{enumerate*}
  \item \textit{Capability}: Malicious adversary controlling up to 5\% of participants,
  \item \textit{Collusion}: Malicious clients coordinate fully, 
  \item \textit{Trusted entities}: Central server assumed honest-but-curious, 
  \item \textit{Acceptable failure}: Accuracy degradation up to 10\% under worst-case poisoning, 
  \item \textit{Evaluation surface}: Evaluated over 100 rounds with IID data across 10,000 users.
\end{enumerate*}

\begin{wrapfigure}{r}
{0.35\textwidth}
\vspace{-7mm}
  \centering  \includegraphics[width=1\linewidth]{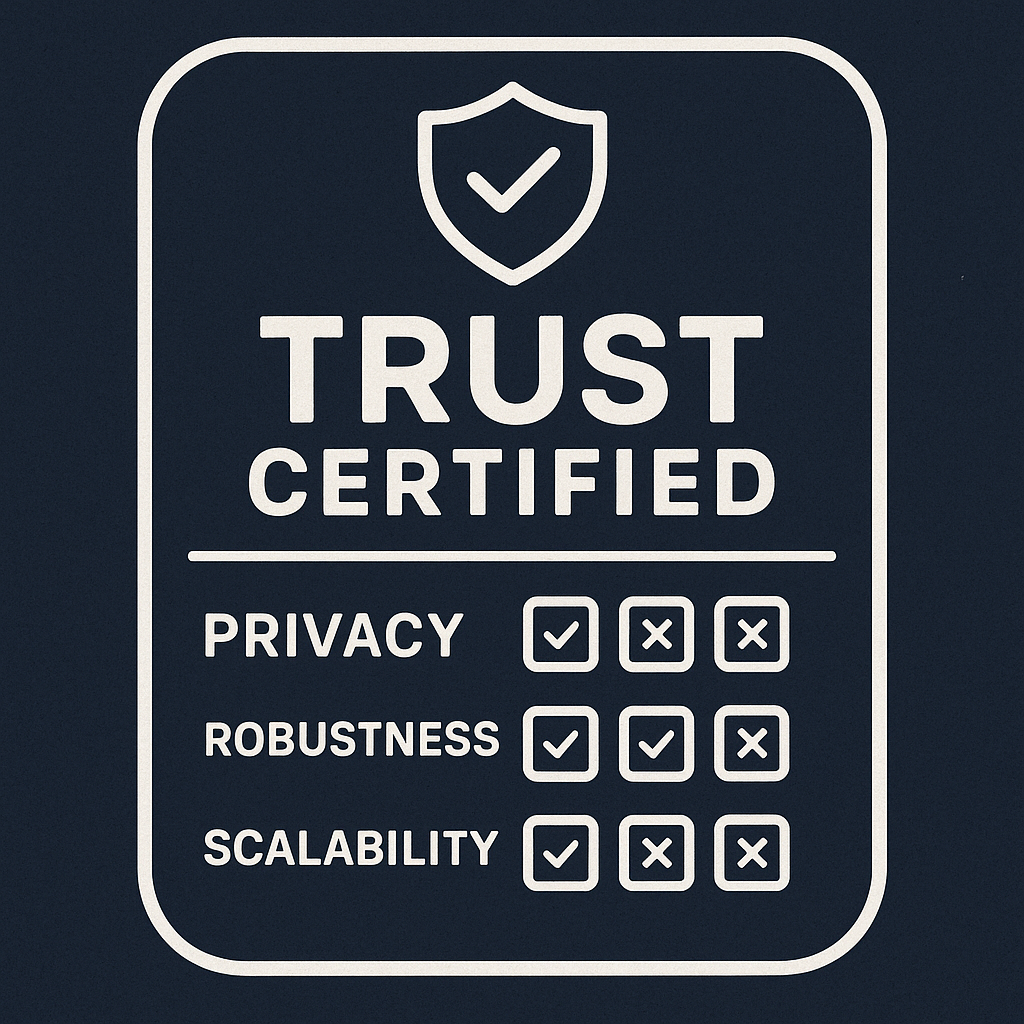}
  \caption{Rather than asserting universal security, the trust certification label should reflect which properties are formally validated and explicit trade-offs.}
  \label{fig:pareto}
  \vspace{-10mm}
\end{wrapfigure}

\paragraph{Certification and trust labels.}
As machine learning systems become integral to critical infrastructure, regulators and users will demand clearer certification of security guarantees. Certification processes must acknowledge the trilemma: no system should claim full privacy, robustness, and efficiency simultaneously without disclosing trade-offs. New trust labeling frameworks could describe systems along three axes, quantifying degrees of robustness, privacy, and efficiency achieved under specific threat models. 
Transparent certification and clear trust labels will be crucial not only for regulatory compliance but also for fostering user trust and guiding informed procurement of ML services.

\section{Alternative Views}
While our main argument emphasizes explicit, context-dependent trade-offs among robustness, privacy, and efficiency, several alternative approaches exist in the literature aiming to simultaneously address all three properties. Below, we briefly review two prominent classes of such alternatives: relaxed formalism approaches, and hybrid cryptographic protocols. We discuss their strengths, limitations, and how each implicitly or explicitly navigates the robustness–privacy–efficiency trilemma.

\subsection{Relaxed DP Formalisms}\label{sec_dp_relax}
A promising strategy softens threat or privacy definitions to shift the trade-off frontier.  
R\'enyi DP and zero-concentrated DP relax $(\varepsilon,\delta)$-DP for tighter composition bounds~\cite{bun2016concentrated,mironov2017renyi}.  
Shuffled-model DP inserts a trusted shuffler to anonymize messages before aggregation, amplifying privacy guarantees so much that the privacy-utility trade-off matches the ideal trusted server scenario (central DP)~\cite{erlingsson2019amplification,cheu2019distributed,girgis2021shuffled}.  
Secret-based DP, e.g., enabling correlated‐noise protocols~\cite{imtiaz2021correlated,sabater2022accurate,allouah2024privacy,allouah2025towards}, leverages pairwise shared randomness to inject lower‐variance noise while preserving both privacy and partial robustness~\cite{sabater2022accurate}, and approaches the central DP privacy-utility trade-off~\cite{allouah2025towards}.  
However, these relaxations necessarily weaken the original guarantees or insert trust assumptions: Rényi DP’s $(\alpha,\varepsilon)$ bounds, for a fixed $\alpha$, do not directly map to DP  as they bound the average (not worst-case) privacy risk, shuffled DP assumes a trusted shuffler, and secret-sharing schemes rely on secure shared randomness among honest participants.  

\textbf{Take-away.} 
In summary, relaxed formalisms can occupy either the \emph{privacy + efficiency} edge (with approximate robustness) or the \emph{robustness + efficiency} edge (with approximate privacy), but they do not achieve all three axes under the strictest definitions.

\subsection{Cryptographic Protocols}
\label{sec_he}
The class of cryptographic approaches aims to protect input confidentiality: e.g., multi-party
computation (MPC) enables an untrusted server to compute aggregates on decentralized data without learning individual records.
This guarantee is \emph{complementary} to our DP objective, which protects the published statistic or model from membership inference.

In contrast to DP, cryptographic primitives such as homomorphic encryption (HE) operate under a weaker yet more practical threat model, assuming that adversaries are computationally bounded.
This relaxation enables the design of protocols that support aggregation and computation directly on encrypted client updates, without requiring noise injection, thereby preserving model accuracy to a greater extent~\cite{phong2018privacy,zhang2020}.
In our context, to address the utility limitations of DP with robustness constraints,
Choffrut et al.~\cite{choffrut2023towards} implement the trimmed mean aggregation~\cite{yin2018byzantine} entirely within the homomorphic encryption domain.
Their approach matches the accuracy of non-private baselines on standard ML benchmarks like MNIST and CIFAR-10.
However, this improved utility comes at the cost of significant computational overhead, primarily due to the complexity of implementing high-dimensional robust aggregation in the encrypted domain.
For example, in their system~\cite{choffrut2023towards}, a single robust aggregation on CIFAR-10 across $n=9$ clients and a model with $d=713,000$ parameters takes over two minutes, which represents a 110$\times$ slowdown compared to a non-robust HE baseline~\cite{zhang2020}, despite the latter using a larger model.
Client subsampling reduces this cost to 39 seconds (3.1\,$\times$ speed-up) with minimal utility loss on CIFAR-10, showing that HE can shift the privacy–utility frontier at the price of compute and latency.
Because HE hides individual updates from an honest-but-curious server, clients may inject less DP noise and still satisfy global privacy
when the final model is released~\cite{kairouz2021distributed}.  
This hybrid approach improves the privacy–utility trade-off while leaving robustness and
compute cost largely unchanged.

General MPC protocols provide simulation-based privacy~\cite{goldreich2004foundations} but typically assume
either a majority of non-colluding servers~\cite{corrigan2017prio,he2020secure,hao21} or incur heavy
communication rounds between all parties~\cite{evans2018pragmatic}, which may be prohibitive in large-scale machine learning.
Thus they trade stronger trust assumptions or higher bandwidth/latency for
noise-free accuracy.
Secure aggregation protocols such as Bonawitz et al.~\cite{bonawitz2017practical} are
light-weight MPC instances now widely deployed in FL, e.g., GBoard.  
Secure aggregation achieves confidentiality against honest-but-curious servers, but
does not entirely address robustness. Indeed, fully malicious participants can still submit
poisoned but correctly masked updates so that statistical defenses, like robust aggregation, remain
necessary.

\textbf{Take-away.}
In summary, cryptography \emph{relaxes} the privacy–utility corner of our trilemma, yet
inevitably moves stress to other axes: computational cost, latency, or
additional trust assumptions.  Robustness must still be enforced by
orthogonal statistical methods.

\section{Conclusion and Future Directions}

The robustness–privacy–efficiency trilemma is a fundamental constraint shaping every large-scale machine learning deployment. Through theoretical bounds, empirical evidence, and practical case studies, we have demonstrated that simultaneously achieving maximal levels of robustness, privacy, and efficiency is infeasible under conventional worst-case threat models.

Recognizing and explicitly navigating this trilemma offers a constructive path forward. We propose three core design principles:

\begin{enumerate}
\item \textbf{Explicit prioritization:} Clearly select two primary objectives—robustness, privacy, or efficiency—tailored to application-specific risks, and deliberately relax the third.
\item \textbf{Joint benchmarking:} Transparently report privacy guarantees (e.g., $(\varepsilon,\delta)$-DP), robustness metrics (e.g., fraction of tolerated adversaries), and efficiency indicators (computational and communication costs) side-by-side for holistic evaluation.
\item \textbf{Transparent certification:} Implement clear and standardized trust labels and certification frameworks that explicitly communicate trade-offs to stakeholders, regulators, and users.
\end{enumerate}

To advance this structured approach, we highlight critical research directions:

\begin{itemize}
\item Develop a comprehensive taxonomy and formal lower bound framework that systematically quantifies how specific relaxations in threat assumptions (e.g., semi-honest intermediaries, bounded adversarial coordination, probabilistic sybil detection) affect achievable trade-offs among robustness, privacy, and efficiency.
\item Investigate and design advanced adaptive algorithms and control mechanisms that dynamically adjust privacy noise levels, robustness thresholds, and computational or communication resource allocation based on real-time detection and forecasting of adversarial behavior and operational constraints.
\item Create rigorous, standardized benchmarks and certification protocols that systematically measure, validate, and transparently communicate robustness–privacy–efficiency trade-offs, facilitating informed decisions by practitioners, policy-makers, and end-users.
\end{itemize}

Arguably, the path to trustworthy ML systems lies not in seeking unattainable perfection but in transparently acknowledging and systematically managing these inevitable trade-offs.

\section*{Acknowledgements}
YA acknowledges support by SNSF  grant 200021\_200477.

\bibliographystyle{plain}
\bibliography{references}

\appendix
\section{Privacy and Communication Efficiency}
\label{sec_priv_comm}
For completeness, we state a trade-off between communication efficiency, privacy, and utility.
We consider the earlier mean estimation estimation task.
Here, we drop the robustness constraint, and instead focus on communication efficiency, in the sense that each user can transmit at most $b$ bits.
Below, we paraphrase a result due to Suresh et al.~\cite{suresh2017distributed}.

\begin{theorem}[Communication-Privacy Lower Bound {\cite{suresh2017distributed}}, paraphrased]
\label{lem:comm-priv}
Consider distributed mean estimation in $\mathbb{R}^{d}$ with $n$ users.  
Any algorithm achieving $(\varepsilon,\delta)$‑local differentially privacy, while each user transmits at most
$b = \mathcal{O}(d)$ bits, must incur error
\begin{equation}
\mathbb{E}\bigl\|\hat{\mu}-\mu\bigr\|_2^2
=
\Omega\left(\frac{d}{n \min\left\{\varepsilon^{2}, b\right\}}\right).
\end{equation}
\end{theorem}
This result shows that any algorithm must trade error or accuracy for communication efficiency. Specifically, decreasing the number $b$ of bits transmitted per user, below a privacy budget-dependent threshold, necessarily increases the best achievable error.
Moreover, this bound is provably tight~\cite{suresh2017distributed}.
Finally, Chen et al.~\cite{chen2020breaking} prove a stronger lower bound in the $\varepsilon \geq 1$ regime, as well as a matching upper bound.

\end{document}